\documentclass[10pt,twocolumn,letterpaper]{article}

\usepackage{iccv}
\usepackage{times}
\usepackage{epsfig}
\usepackage{graphicx}
\usepackage{amsmath}
\usepackage{amssymb}
\usepackage{multirow}
\usepackage{booktabs}
\usepackage{pifont}
\usepackage[accsupp]{axessibility} 

\usepackage[pagebackref=true,breaklinks=true,letterpaper=true,colorlinks,bookmarks=false]{hyperref}

\iccvfinalcopy 

\def\iccvPaperID{1476} 

\begin{document}

\title{Not Every Side Is Equal: Localization Uncertainty Estimation for Semi-Supervised 3D Object Detection}

\author{Chuxin Wang$^1$, Wenfei Yang$^1$, Tianzhu Zhang$^{1,2,}$\footnotemark[1]~\\
{$^1$University of Science and Technology of China, $^2$Deep Space Exploration Lab}\\
{\tt\small {wcx0602@mail.ustc.edu.cn}, {yangwf@mail.ustc.edu.cn}, {tzzhang@ustc.edu.cn}}\\
}


\maketitle

\renewcommand{\thefootnote}{\fnsymbol{footnote}}
\footnotetext[1]{Corresponding Author}


\begin{abstract}

Semi-supervised 3D object detection from point cloud aims to train a detector with a small number of labeled data and a large number of unlabeled data.
The core of existing methods lies in how to select high-quality pseudo-labels using the designed quality evaluation criterion. 
However, these methods treat each pseudo bounding box as a whole and assign equal importance to each side during training, which is detrimental to model performance due to many sides having poor localization quality. 
Besides, existing methods filter out a large number of low-quality pseudo-labels, which also contain some correct regression values that can help with model training.
To address the above issues, we propose a side-aware framework for semi-supervised 3D object detection consisting of three key designs: a 3D bounding box parameterization method, an uncertainty estimation module, and a pseudo-label selection strategy.
These modules work together to explicitly estimate the localization quality of each side and assign different levels of importance during the training phase.
Extensive experiment results demonstrate that the proposed method can consistently outperform baseline models under different scenes and evaluation criteria. 
Moreover, our method achieves state-of-the-art performance on three datasets with different labeled ratios.
%

\end{abstract}

\section{Introduction}

The goal of 3D object detection is to estimate oriented 3D boundary boxes as well as category labels of objects from a point cloud.
Due to its significant potential applications in autonomous driving~\cite{geiger2012we,gomez2016pl,arnold2019survey},
many fully supervised 3D object detection methods have been proposed over the past few decades~\cite{shi2020pv, rukhovich2021fcaf3d},
leading to remarkable progress in this area.
However, these methods rely heavily  on a large amount of carefully annotated 3D scene data,
which is expensive and time-consuming to collect.
\label{sec:intro}

\begin{figure}[!t]
    \begin{center}
        \includegraphics[width=0.47\textwidth]{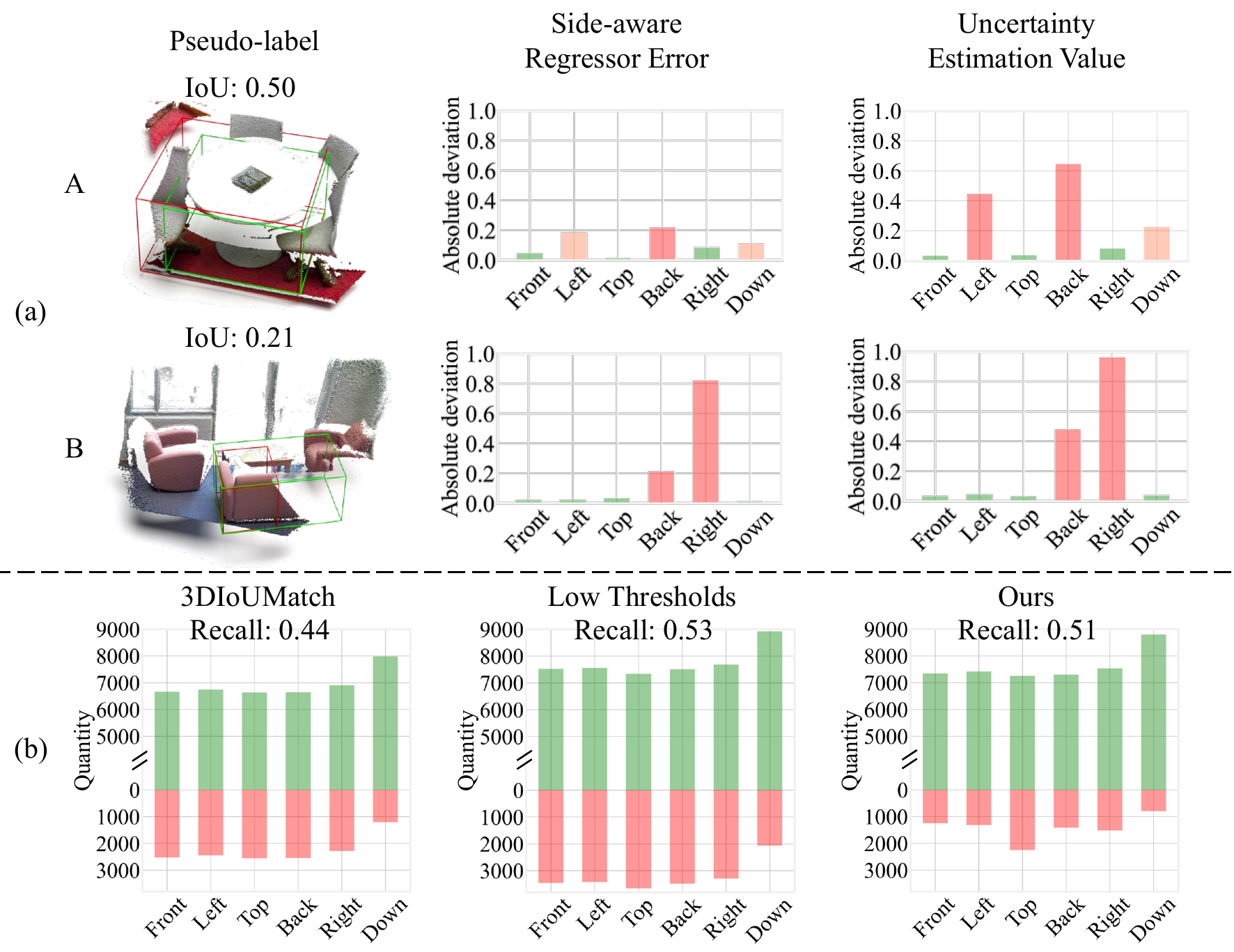}
        \caption{\textbf{The motivation of this paper.} 
        (a): Red represents pseudo-labels and green represents ground-truth. Pseudo-labels with high IoU may be incorrect on some sides, while pseudo-labels with low IoU may be correct on some sides. Global scores like IoU are insufficient for pseudo-label selection. 
        (b): The side quality statistics of different methods for 300 scenes. The green bars indicate the number of sides with error less than 0.1, while the red bars indicate the number of sides with error greater than 0.1. The recall indicates the proportion of ground-truth that are covered by the pseudo-labels. The proposed method can effectively select sides with high quality for model training.}
        \label{tissue}
    \end{center}
\end{figure}

To reduce the high annotation cost associated with fully supervised 3D object detection methods,
semi-supervised methods~\cite{tang2019transferable,cvpr2020sess,icip2021semi, cvpr20213dioumatch, zhang2022atf} that use a combination of few labeled data and a large amount of unlabeled data for model training have gained increasing attention.
Existing semi-supervised methods can be broadly divided into two categories: consistency-based methods~\cite{cvpr2020sess} and pseudo-label based methods~\cite{icip2021semi, cvpr20213dioumatch, zhang2022atf}.
For consistency-based methods, the core idea is to encourage the prediction on data with different augmentations to be consistent. 
For example, SESS~\cite{cvpr2020sess} adopts the mean-teacher framework and utilizes three consistency losses between the student and teacher model to enforce the consensus of object locations, semantic categories and sizes. 
However, the consistency constraint may lead to suboptimal results due to significant noise in model predictions.
On the other hand, pseudo-label based methods aim to select high-quality pseudo-labels from model predictions of unlabeled data and then use them in combination with labeled data for model training.
For example, HPCA~\cite{icip2021semi}  introduces the idea of adaptive class confidence selection to select pseudo-labels with confidence above the threshold, while 3DIoUMatch~\cite{cvpr20213dioumatch} equips the detector with 3D Intersection over Union (IoU) estimation module to estimate the global localization quality and proposes lower-half suppression strategy for pseudo-label filtering. 

Despite the success of pseudo-label based methods, they all use global metric scores (IoU, classification confidence, and voting score, et al.) to select pseudo-labels. 
However, pseudo-labels with high global metric scores  may not cover each side of the object well, while pseudo-labels with low global metric scores may provide correct predictions for some object  sides.
As shown in Figure~\ref{tissue} (a), the IoU of pseudo-label A is high, but significant errors exist on the left and back sides. 
Using these noisy predictions to supervise the student model is harmful. 
On the contrary, although the IoU of pseudo-label B is relatively low, the predictions of four sides (front, left, top and down sides) are accurate, which is valuable for model training but discarded in existing methods.

To deal with the above problem, an intuitive way is to estimate the localization quality of each side and treat different sides with different importance  for model training. 
To achieve this, two key issues need to be considered:  
1) \textit{How to estimate the localization quality of each side?}
Existing methods~\cite{icip2021semi, cvpr20213dioumatch, zhang2022atf} treat the boundary box prediction as a whole and usually use global metrics to supervised model training, 
which cannot reflect the localization quality of each side. 
%
2) \textit{how to select pseudo-labels based on the side localization quality for model training?}
An effective pseudo-label selection strategy should suppress the interference of sides with low quality while retaining sides with high quality to supervise model training. 

Motivated by the above discussion, we propose a novel side-aware method for semi-supervised 3D object detection,
%
which consists of a new 3D boundary box parameterization strategy, a quality estimation strategy, and a pseudo-label filtering strategy.
In the 3D boundary box parameterization strategy, 
we decouple the localization task by dividing the object boundary box into six independent sides and predict the position of each side as a probability distribution over discrete values.
The distribution can determine the location of each side and can be used in subsequent modules for localization quality estimation. 
In the quality estimation strategy, 
we design a side-aware uncertainty estimation module to evaluate the quality of each side. 
It takes  the predicted distribution properties and side-aware geometric features as input,
and learns to assess the quality of each side under the guidance of an uncertainty regression loss. 
As shown in Figure~\ref{tissue}(a), the proposed strategy can assign higher uncertainty to sides with lower localization quality.
In the pseudo-label filtering strategy, 
we use class-specific dynamic thresholds and a soft weight assignment strategy 
to assign higher importance to sides with higher quality, 
which can suppress the interference of low-quality  sides while fully utilizing sides with higher quality in the pseudo-labels.
As shown in Figure~\ref{tissue} (b), 3DIoUMatch~\cite{cvpr20213dioumatch} filters out low-quality pseudo-labels by setting a high threshold, which discards a large number of correct pseudo-labels.
However, a low threshold can improve the recall ratio, and it also introduces many incorrect pseudo-labels. 
Our method can suppress the interference of incorrect sides and retain as many correct pseudo-labels as possible for model training.

%

%
In summary, the contributions of this paper are as follows:
%
%
(1) We introduce a novel side-aware model for semi-supervised 3D object detection, which includes a 3D bounding box parameterization method, an uncertainty estimation module, and a pseudo-label selection strategy. The proposed method allows for explicit estimation of the localization quality of each side and can assign different levels of importance to each side during the training phase.
(2) Extensive experimental results show that the proposed method can consistently improve the performance of baseline detectors under different scenes and evaluation criteria,
verifying the effectiveness of our model.
Moreover, the proposed method can achieve new state-of-the-art performance on 3 datasets, including 2 indoor datasets and 1 outdoor dataset.

\section{Related Work}

%
In this section, we briefly review semi-supervised object detection and uncertainty estimation methods. 
\subsection{Semi-supervised object detection.}
%
The great progress of fully supervised object detectors~\cite{he2019bounding, li2020generalized, qi2019deep, liu2021group, rukhovich2021fcaf3d, shi2019pointrcnn, zhang2020h3dnet} relies on a large amount of carefully annotated data,
the semi-supervised methods~\cite{cvpr2020sess, cvpr20213dioumatch, icip2021semi,li2020improving, tang2021proposal, xu2021end, liu2021unbiased} that utilize few labeled data and large amount of unlabeled data for model training have been attracting more and more attention.
In general, these methods can be divided into  consistency based methods~\cite{cvpr2020sess,jeong2019consistency, tang2021proposal} and pseudo-label based methods~\cite{cvpr20213dioumatch,sohn2020simple,radosavovic2018data,zoph2020rethinking,li2020improving, xu2021end, liu2021unbiased}.
In this paper, we mainly focus on the pseudo-label based methods.
In the 2D area, STAC~\cite{sohn2020simple} proposes to train a detector on labeled data first,
and then generates pseudo-labels for unlabeled data, which are then used with labeled data to re-train the detector.
In~\cite{radosavovic2018data} and~\cite{zoph2020rethinking},
the ensembles  of predictions on data with different augmentations are used to form the pseudo-labels for unlabeled images.
In SoftTeacher~\cite{xu2021end}, they propose the first end-to-end pseudo-labeling framework by utilizing a teacher model to avoid the complicated multi-stage training process.
%
In the 3D area,
SESS~\cite{cvpr2020sess} is the first work for semi-supervised point-based 3D object detection,
which consists of a student model and an EMA teacher model.
%
%
By constraining the outputs of these two models to be consistent, it can effectively learn from these unlabeled data.
However, it is suboptimal to enforce all predictions of the student and teacher model to be consistent,
because there are many false predictions.
To mitigate this issue, 3DIoUMatch~\cite{cvpr20213dioumatch} introduces the confidence-based filtering strategy and IoU prediction strategy to select pseudo-labels with high quality predicted by the teacher model.
%
%
In line with this work, the contemporary work HPCA~\cite{icip2021semi} design an adaptive class confidence selection scheme for pseudo-label filtering,
which is inspired by the FlexMatch~\cite{nips2021flexmatch} strategy for semi-supervised image classification.
In this paper, we focus on localization quality estimation.
Different from existing methods that treat each bounding box as a whole and assign a global quality score for it,
we aim to evaluate the localization quality of each side and treat them with different importance during the training stage.
\subsection{Uncertainty estimation.}
%
Uncertainty estimation aims to produce a measure of confidence for model predictions,
which plays an important role in many AI systems, such as autonomous driving.
Traditional deep neural networks are deterministic models,
and the bayesian deep learning makes it possible to estimate the uncertainty of deep model predictions~\cite{cvpr2021masksembles}.
In~\cite{nips2017uncertainties},
a bayesian deep learning framework is proposed to estimate the
aleatoric uncertainty about data and epistemic uncertainty about model, respectively.
Inspired by this work, many subsequent works have been proposed for various computer vision tasks,
such as object detection~\cite{iccv2019gaussian, meyer2019lasernet}, medical image segmentation~\cite{tmi2021inconsistency}, person re-identification~\cite{aaai2021exploiting}.
In~\cite{he2019bounding} and~\cite{iccv2019gaussian},
they introduce the idea of uncertainty estimation into the object detection area by modeling the bounding box
coordinates as the gaussian parameters.
Different from these works that treat the bounding box as gaussian distribution,
Li et al.~\cite{li2020generalized} propose to represent the bounding box locations as arbitrary distribution by learning a discretized probability vector over the continuous space.
For 3D object detection, a probabilistic detector is proposed in~\cite{meyer2019lasernet} to quantify uncertainty for LiDAR point cloud vehicle detection.
Based on this work,~\cite{Meyer2020Uncertainty} improve the ability to learn the probability distribution by considering the potential noise in the ground-truth labeled data.
In~\cite{Lu2021Geometry} and~\cite{Pan2020Uncertainty}, geometry infromation is used to predict the uncertainty of the detection bounding box.
Different from previous methods, we design an uncertainty estimation method for each side, which incorporates the spatial distribution properties of the sides and the nearby geometric properties.
%

\section{Method}
\begin{figure*}[!t]
    \begin{center}
        \includegraphics[width=0.95\textwidth]{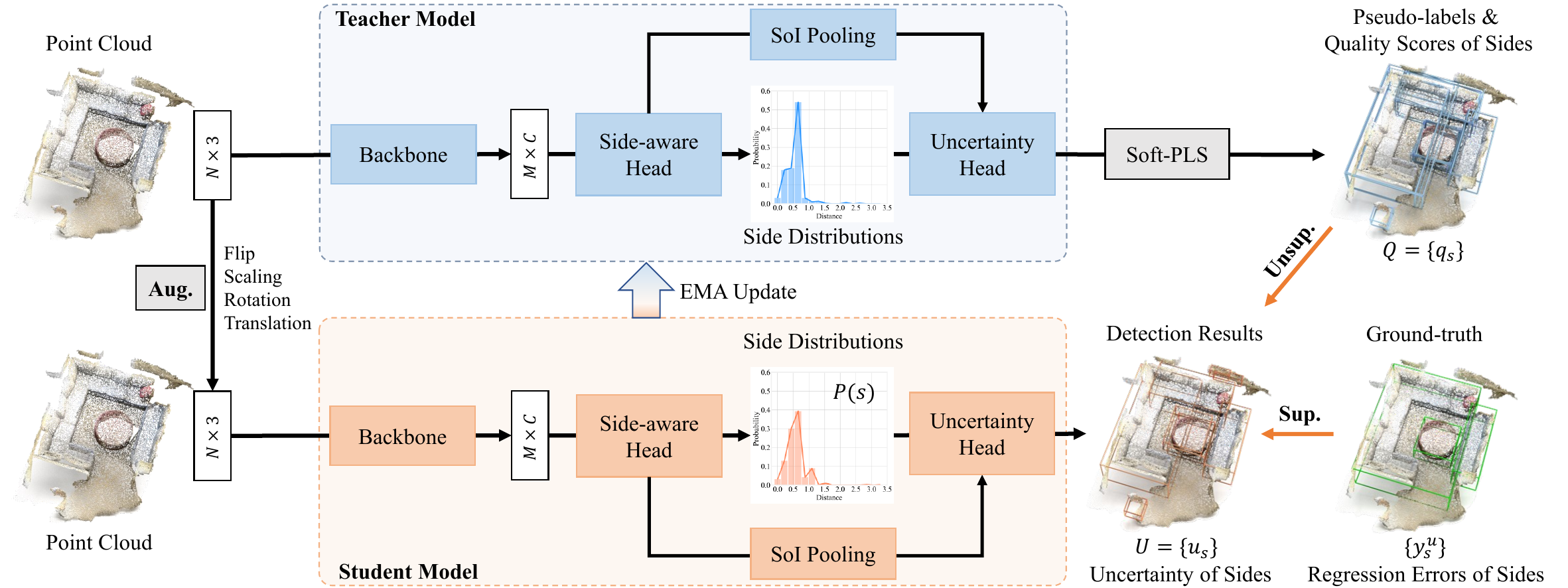}
        \caption{\textbf{The overview of the proposed side-aware framework for semi-supervised 3D object detection.}
        Our method is based on the mean-teacher framework.
        We start by augmenting the point cloud data and feeding it into the student model to obtain both the detection result and the  uncertainty of the sides. 
        For labeled data, we directly use the ground-truth to constrain the predicted results. 
        For unlabeled data, we apply the soft-PLS method to filter the predicted results of the teacher model to generate pseudo labels,
        which are used to supervise the student model.
        %
        %
        To update the parameters of the teacher model, we employ the Exponential Moving Average (EMA) strategy.
        }
        \label{framework}
    \end{center}
\end{figure*}

%

\subsection{Overview}
\label{Overview}
%
%
%
%
Figure~\ref{framework} illustrates the pipeline of our approach.
Given a point cloud, proposal features are obtained using a backbone model. 
We design a side-aware detection head, which can obtain the distribution of the six sides of candidate objects in 3D space. 
By integrating the spatial distribution of each side, we are able to determine the position of each side. 
Then, we use Side of Interest Pooling (SoI Pooling) to extract the geometric features of each side and fuse them with distribution properties. 
The fused features are then input into the uncertainty estimation head to estimate the uncertainty of the sides. 
For labeled data, we directly use the ground-truth to constrain the predictions. 
For unlabeled data, we employ soft Pseudo-Label Selection (soft-PLS) to filter the predicted results of the teacher model, thereby obtaining pseudo-labels and the quality score of each regression value. 
Then we use the pseudo-labels to supervise the student model.

\subsection{Side-aware 3D Object Detection}
\label{Side-aware 3D Object Detection}
Existing 3D object detection methods~\cite{shi2020pv,cvpr20213dioumatch} adopt an extra IoU prediction branch to estimate the global localization quality. 
%
%
%
As explained in Section~\ref{sec:intro}, it is important to evaluate the quality of each side.
To achieve this,
we take inspiration from~\cite{he2019bounding} and design a side-aware model dedicated to semi-supervised 3D object detection, which includes three specific designs. The details are as follows.
%
%
%
%

%
\noindent\textbf{Side-aware Bounding Box Parameterization.}
%
%
%
For semi-supervised 3D object detection task, decoupling the bounding box prediction into independent side prediction is helpful for the uncertainty evaluation of the sides. 
Therefore, we design a side-aware parameterization method to represent the bounding box. 
Specifically, given the features and locations of candidate points, instead of predicting the offsets to the center and the sizes of the object, we predict the distance from the candidate points to each side.
Based on the observation that predicted probability distributions can measure the uncertainty, we modify the bounding box parameterization from a deterministic way to probabilistic way. 
For convenience, we denote the top, down, left, right, front and back sides of a bounding box as $s \in B=\{t, d, l, r, f, b\}$.
%
%
We aim to predict a probabilistic distribution for $s$.
Specifically, denote the possible range  of $s$ as $[s_{min}, s_{max}]$,
the model predicts a probability distribution $P(s)$ defined over the interval.
For convenience of implementations,
we discretize the continuous distribution by dividing the interval $[s_{min}, s_{max}]$ into $N$ bins $(s_0, s_1, \cdots, s_{N-1})$,
and the expected value $\hat{s}$ of $s$ can be calculated as Equation~\eqref{yhat_d}.
\begin{equation}
    \label{yhat_d}
    \hat{s} = \sum_{i=0}^{N-1} P(s=s_i) s_i.
\end{equation}
In previous 3D object detection methods~\cite{qi2019deep, yang20203dssd, liu2021group},
the bounding box center-specific and size-specific regression loss are commonly used for bounding box regression.
%
This formulation is not suitable to the side-aware design,
thus we use the rotated IoU loss~\cite{zhou2019iou} and the side-based smooth L1 loss in the side-aware network.
\begin{equation}
    \label{loss_3d_2}
    \mathcal{L}_{box} = \mathcal{L}_{iou}(B) + \sum_{s \in B}\mathcal{L}_{reg}(s).
\end{equation}
The side-based smooth L1 loss $\mathcal{L}_{reg}(s)$ focuses on the local localization of each side,
which is conducive to the subsequent uncertainty prediction.
The rotated IoU loss $\mathcal{L}_{iou}(B)$ focuses on the global localization of the bounding box,
which is robust to shape and scale variations.

\begin{figure}[!t]
    \begin{center}
        \includegraphics[width=0.45\textwidth]{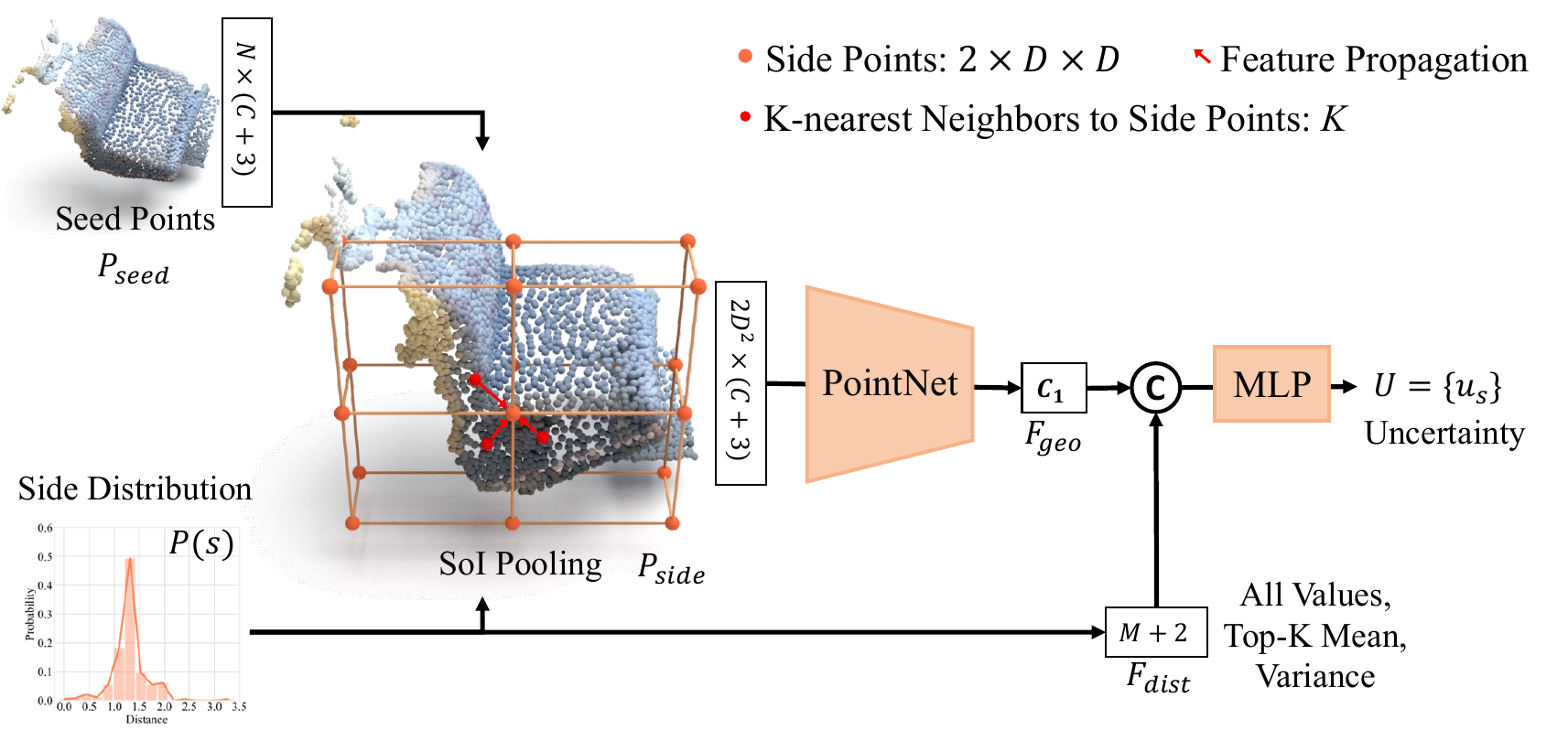}
        \caption{\textbf{The Side-aware Uncertainty Estimation Module.} The geometric features and distribution properties are fused together to estimate the uncertainty of each side.
        }
        \label{uncer_module}
    \end{center}
\end{figure}

\noindent\textbf{Uncertainty Estimation Module.}
Given the predicted spatial distribution of the six sides, one naive approach is to directly utilize statistical measures such as variance to assess the uncertainty of localization results.
However, this approach has certain limitations. 
Firstly, the relationship between distribution properties and the localization quality of sides is complex, and cannot be accurately measured by simple statistics.
Secondly, the side-aware localization quality of objects with occluded structures is challenging to determine based on distribution properties.
Therefore, we designed the uncertainty estimation module, as shown in Figure~\ref{uncer_module}, 
which combines the distribution properties and geometric features of the side to predict its localization uncertainty through MLP.
Specifically, we first generate a series of side points $P_{side}$ near each side. 
For each side point, we find its k-nearest neighbors and apply a distance-weighted interpolation for feature propagation from the seed points $P_{seed}$ to the side point. 
Then, we input all the side points into a PointNet to obtain the geometric features $F_{geo}\in \mathcal{R}^{C_1}$. 
Simultaneously, we concatenate the distribution of the side and corresponding statistics (Top-$k$ value mean and variance) to obtain the distribution properties $F_{dist}\in \mathcal{R}^{M+2}$. 
Finally, we fuse the geometric features and distribution properties into a MLP to obtain the uncertainty measure $U=\{u_s|s \in B\}$ of sides, 
as follows: 
\begin{equation}
    \label{uncer}
    u_s = \operatorname{Sigmoid}(\operatorname{MLP}(\operatorname{Cat}(F_{geo}, F_{dist}))).
\end{equation}

\noindent\textbf{Uncertainty Regression Loss.}
To guide the training of the uncertainty estimation module,
we introduce the uncertainty regression loss into our case. 
We ultize the absolute deviation of the predicted side location $\hat{s}$ and ground-truth $y_s$ to compute the uncertainty label $y^s_u$, as follows:
\begin{equation}
    \label{side_label}
    y^s_u = \operatorname{MIN}(\alpha_1 | y_s - \hat{s} |, 1.0), 
\end{equation}
where $\alpha_1$ is the scaling factor, which is set to be 4 in this paper. 
%
%
The uncertainty regression loss can now be computed as the mean square error between $y^s_u$ and $u_s$:
\begin{equation}
    \label{loss_uncer}
    \mathcal{L}_{uncer.} = \frac{1}{|U|}(y^s_u - u_s)^2
\end{equation}
where $U$ is the uncertainty estimation $U=\{u_s|s \in B\}$ of each side of the object.
With this loss constraint,
the uncertainty estimation module can learn from labeled data to measure the localization uncertainty of each side.
%

\subsection{Soft Pseudo-Label Selection}
\label{Uncertainty Confidence Selection}
In the semi-supervised setting, the performance of the detector depends heavily on the quality of the pseudo-labels.
Thus, we propose the Soft Pseudo-Label Selection (soft-PLS), which consists of three components: category-specific filter, IoU-guided NMS with low-half strategy, side-aware weight assignment. 
For category-specific filter, we select pseudo-labels by jointly considering the classification score, objectness score and IoU score. 
As pointed out by previous works~\cite{icip2021semi},
using fixed thresholds for all categories is suboptimal because it ignores the different learning difficulties of different categories.
Thus we follow FlexMatch~\cite{nips2021flexmatch} and use category-specific thresholds to filter pseudo-labels. 
Denote the threshold for classification score, objectness score and IoU score as $\tau_{cls}$, $\tau_{obj}$  and $\tau_{iou}$,
Specifically, we calculate a category-specific scale factor $\gamma_t(c)$ at time stamp $t$ (refer to the supplementary material for detailed procedures), 
and the adaptive threshold $\tau_t(c)$ for category $c$ can be calculated as follows:
\begin{equation}
 \label{tau_c}
  \tau_t(c) = \tau_{min} + (\tau_{max}- \tau_{min}) \gamma_t(c),
\end{equation}
where $\tau_{min}$ and $\tau_{max}$ controls the range of the adaptive threshold.
Since $\tau_{obj}$ is class-agnostic, we only apply this adaptive threshold strategy on  $\tau_{cls}$ and $\tau_{iou}$.
After obtaining pseudo-labels through the category-specific thresholds, we need to suppress noise in pseudo-labels caused by duplicated bounding box predictions.
Therefore, we utilize the IoU-guided non-maximal suppression with low-half keeping strategy~\cite{cvpr20213dioumatch} to eliminate redundant pseudo-labels.
Specifically, for a bunch of highly-overlapped pseudo-labels, we only discards half of the proposals with lower predicted IoU. 
Although better pseudo-labels can be obtained through the above modules, there are many pseudo-labels with poor localization quality on several sides, which is detrimental to model training. 
To mitigate this issue,
we first evaluate the localization quality of each side with the estimated uncertainty $u_s$.
Specifically, we first compute the quality score $Q = \{q_s|s \in B\}$ of each side as follows:
\begin{equation}
    \label{quality_score}
    q_s = e^{-\alpha_2  u_s}, 
\end{equation}
where $\alpha_2$ is a scaling value. 
When supervising the student model, we use the quality scores $Q$ of the pseudo-label to weight the loss function as follows:
\begin{equation}
    \label{weighted_function}
    \mathcal{L}_{box} = q_B \mathcal{L}_{iou}(B) + \sum_{s \in B}(q_s \mathcal{L}_{reg}(s)), 
\end{equation}
Where $q_B$ is the mean value of $q_s$ and reflects the global localization quality of the bounding box.
In this way, we reduce the interference of poorly localized sides in model training.
In summary, we obtain high-quality pseudo-labels and corresponding quality scores of the sides by the proposed soft-PLS.

\subsection{Model Training}
\label{SSL for Semi-supervised Object Detection}
In this section, we introduce how to train the model under the semi-supervised setting.
Our approach follows the Mean Teacher paradigm~\cite{tarvainen2017mean}.
%
%
In the pre-training stage, we train the side-aware model on the labeled dataset $\{X^l, Y^l\}$.
The training loss is the same as the original detector except that the regression loss is replaced with the side-aware regression loss in Equation~\eqref{loss_3d_2} and add the uncertainty regression loss in Equation~\eqref{loss_uncer}.
Once converged, we clone the model to create a pair of student and teacher models.
In the semi-supervised training stage,
labeled data $\{X^l, Y^l\}$ and unlabeled data $\{X^u\}$ are randomly sampled according to a predefined sampling ratio $r$ in each training iteration.
For unlabeled data,
we apply two different augmentations to the input of the teacher model and student model, respectively.
%
%
The teacher model outputs the pseudo-labels $\{\hat{Y}^u\}$ with quality scores of the sides $Q$,
which are then used together with the labeled data $\{X^l, Y^l\}$ to train the student model.
For the supervised loss, $\mathcal{L}(X^l, Y^l)$ is computed in the same way as the pre-training stage.
For the unsupervised loss, $\mathcal{L}(X^u, \hat{Y}^u)$ is defined as the sum of Equation~\eqref{weighted_function} and the classification loss in the baseline detector.
The overall loss is defined as the weighted sum of the supervised loss and unsupervised loss as follows:
\begin{equation}
    \label{loss_ssl}
    \mathcal{L} = \mathcal{L}(X^l, Y^l) + \beta \mathcal{L}(X^u, \hat{Y}^u),
\end{equation}
where $\mathcal{L}(X^l, Y^l)$ and $\mathcal{L}(X^u, \hat{Y}^u)$ denote the loss on labeled data and unlabeled data, respectively.
$\beta$ controls the contribution of the unsupervised loss.
During the training stage,
the teacher model is updated by the student model with an Exponential Moving Average (EMA) strategy~\cite{tarvainen2017mean}.
%

\subsection{Discussions}

Probabilistic models are also used in GFLV2~\cite{li2021generalizedv2} and LaserNet~\cite{meyer2019lasernet} for bounding box quality estimation, and the differences are discussed as follows:
(1) In GFLV2, the main goal is to predict the global localization quality based on the spatial distribution of 2D bounding boxes. 
Specifically, GFLV2 employs a distribution-guided quality predictor to estimate a joint representation of classification and IoU. 
while in our method, the main goal is to estimate the quality score of each side individually and treat them with different importance during the training stage. 
We utilize the distribution properties and the geometric features of each side to estimate the uncertancy of each side individually and propose the side-aware uncertainty loss to guide the training of the uncertainty module.
(2) LaserNet~\cite{meyer2019lasernet} models the global uncertainty of the bounding box based on pixel predictions within the same cluster. 
Specifically, LaserNet predicts the shared variance of the bounding box corners for each pixel and use mean shift clustering to obtain the bounding box with the distribution variance.
Different from LaserNet, our approach predicts the spatial distribution of each side of the bounding box and evaluates the uncertainty of each side separately. 
%

\section{Experiments}
%
\subsection{Experimental Setup}
\label{sec:expset}
\textbf{Dataset.}
We evaluate the proposed method on two indoor datasets and one outdoor dataset,
including ScanNet~\cite{dai2017scannet}, SUNRGB-D~\cite{song2015sun} and KITTI~\cite{geiger2012we}.
For the indoor datasets, \textbf{ScanNet}~\cite{dai2017scannet} dataset contains 1.2K training samples and 312 validation samples annotated with per-point instances, semantic labels, and axis-aligned 3D bounding boxes belonging to 18 categories.
\textbf{SUNRGB-D}~\cite{song2015sun} dataset consists of ~5K single-view indoor RGB-D images annotated with per-point semantic labels, and oriented 3D bounding boxes belonging to 37 categories.
For the outdoor dataset, \textbf{KITTI}~\cite{geiger2012we} dataset contains 7481 outdoor scenes for training and 7518 scenes for testing, and the training samples are generally divided into a training split of 3712 samples and a validation split of 3769 samples.
\textbf{For all datasets,} we follow~\cite{cvpr2020sess}to evaluate on different proportions of labeled data randomly sampled from all the training data.
We keep the remaining data as unlabeled data for training in our semi-supervised framework.
As for 100\% labeled data, we simply make a copy of the full dataset as unlabeled data.
\textbf{Evaluation Metrics.}
For the indoor datasets, following the standard evaluation protocol~\cite{qi2019deep}, we evaluate the  performance on the Val set with the mean Average Precision (mAP) values under intersection over union (IoU) thresholds 0.25 and 0.5,
denoted as $\mbox{mAP}_{25}$ and $\mbox{mAP}_{50}$.
For the outdoor dataset, We follow~\cite{shi2020pv} for data pre-processing and report the mAP with 40 recall positions, with a rotated IoU threshold 0.7, 0.5, 0.5 for the car, pedestrian, and cyclist categories, respectively.
Due to the randomness of the data splits,
we report the results as mean $\pm$ standard deviation across 3 runs under different random data splits.

\textbf{Implementation Details.} 
%
%
For the indoor datasets, we use VoteNet~\cite{qi2019deep} as the baseline model to be in line with previous works~\cite{cvpr2020sess,cvpr20213dioumatch}. 
We analysis the sizes of all objects in the indoor datasets and set the ranges of front, back, left and right sides as [0, 3.5], while the ranges of top and bottom sides are [0, 2.0]. 
In addition, we split the range into $N=32$ bins. 
We consider a side with an error greater than 0.25 to be unreliable, thus we set $\alpha_1=4$ in Equation~\eqref{side_label} and $\alpha_2=5$ in Equation~\eqref{quality_score}.
For pseudo-label selection, the objectness score threshold is set to be $\tau_{obj}=0.8$,
the minimum threshold $\tau_{min}$ for the classification score and IoU score are set to be 0.7 and 0.15,
and the maximum threshold $\tau_{max}$ is set to be 0.9 and 0.25.
For the outdoor dataset, we use PV-RCNN~\cite{shi2020pv} as the baseline model. 
Different from VoteNet~\cite{qi2019deep}, PV-RCNN is a two-stage detector, where a region proposal network (RPN) is used to generate candidate proposals.
The objectness score is predicted by the RPN and used to pick the top-100 proposals for the Region of Interest (RoI) head in the second stage.
In order to maintain the main structure of PV-RCNN, we modify the RoI head as the side-aware head and set the ranges of front and back sides as [0, 0.4], while the ranges of left, right, top and bottom sides are [0, 0.3]. 
Since there are only three categories in this dataset, 
we use fixed IoU thresholds $\tau_{car}=0.7,\tau_{ped}=0.3,\tau_{cls}=0.15$ in the category-specific filter of soft-PLS.
The remaining hyperparameters are consistent with VoteNet. 
Training details please refer to the supplementary material. 

%
\subsection{Comparison on Indoor Datasets}
\begin{table*}[!t]
  \begin{center}
      \setlength\tabcolsep{5pt}
      \caption{\textbf{Results on ScanNet Val dataset under different ratios of labeled data.} Results are reported as mean $\pm$ standard deviation across 3 runs with random data splits, * represents the re-implemented results on ScanNet 50\% labeled data.}
      \small
      \label{table:scannet}
      \begin{tabular}{c|c|c|c|c|c|c|c|c|c|c}
          \toprule
         \multirow{2}*{Model} & \multicolumn{2}{c|}{5\%} & \multicolumn{2}{c|}{10\%} & \multicolumn{2}{c|}{20\%} & \multicolumn{2}{c|}{50\%$^*$} & \multicolumn{2}{c}{100\%}                   \\
          \cline{2-11}
           ~& $\mbox{mAP}_{25}$& $\mbox{mAP}_{50}$ & $\mbox{mAP}_{25}$ & $\mbox{mAP}_{50}$ & $\mbox{mAP}_{25}$ & $\mbox{mAP}_{50}$   & $\mbox{mAP}_{25}$   & $\mbox{mAP}_{50}$   & $\mbox{mAP}_{25}$   & $\mbox{mAP}_{50}$   \\
          \midrule
            VoteNet~\cite{qi2019deep}       &27.9$\pm$0.5&10.8$\pm$0.6&36.9$\pm$1.6&18.2$\pm$1.0&46.9$\pm$1.9&27.5$\pm$1.2&56.1$\pm$1.1&36.5$\pm$0.6&57.8&36.0\\
            SESS~\cite{cvpr2020sess}          &32.0$\pm$0.7&14.4$\pm$0.7&39.5$\pm$1.8&19.8$\pm$1.3&49.6$\pm$1.1&29.0$\pm$1.0&57.2$\pm$1.2&37.7$\pm$0.7&61.3&39.0\\
            3DIoU~\cite{cvpr20213dioumatch}         &40.0$\pm$0.9&22.5$\pm$0.5&47.2$\pm$0.4&28.3$\pm$1.5&52.8$\pm$1.2&35.2$\pm$1.1&59.8$\pm$0.7&41.2$\pm$0.5&62.9&42.1\\
          \textbf{Ours} &\textbf{40.5$\pm$1.1}&\textbf{23.8$\pm$0.8}&\textbf{48.8$\pm$0.9}&\textbf{31.1$\pm$1.1}&\textbf{54.5$\pm$0.8}&\textbf{37.3$\pm$0.5}&\textbf{61.5$\pm$1.4}&\textbf{43.1$\pm$0.8}&\textbf{63.8}&\textbf{44.1}\\
          \bottomrule
      \end{tabular}
  \end{center}
\end{table*}

\begin{table*}[!t]
  \begin{center}
      \setlength\tabcolsep{5pt}
      \caption{\textbf{Results on SUNRGB-D Val dataset under different ratios of labeled data.} Results are reported as mean $\pm$ standard deviation across 3 runs with random data splits, * represents the re-implemented results on SUNRGB-D 50\% labeled data.}
      \small
      \label{table:sunrgbd}
      \begin{tabular}{c|c|c|c|c|c|c|c|c|c|c}
          \toprule
         \multirow{2}*{Model} & \multicolumn{2}{c|}{5\%} & \multicolumn{2}{c|}{10\%} & \multicolumn{2}{c|}{20\%} & \multicolumn{2}{c|}{50\%$^*$} & \multicolumn{2}{c}{100\%}                   \\
          \cline{2-11}
           ~& $\mbox{mAP}_{25}$& $\mbox{mAP}_{50}$ & $\mbox{mAP}_{25}$ & $\mbox{mAP}_{50}$ & $\mbox{mAP}_{25}$ & $\mbox{mAP}_{50}$   & $\mbox{mAP}_{25}$   & $\mbox{mAP}_{50}$   & $\mbox{mAP}_{25}$   & $\mbox{mAP}_{50}$   \\
          \midrule
            VoteNet~\cite{qi2019deep}       &29.9$\pm$1.5&10.5$\pm$0.5&38.9$\pm$0.8&17.2$\pm$1.3&45.7$\pm$0.6&22.5$\pm$0.8&55.3$\pm$1.1&31.9$\pm$0.8&58.0&33.4     \\
            SESS~\cite{cvpr2020sess}           &34.2$\pm$2.0&13.1$\pm$1.0&42.1$\pm$1.1&20.9$\pm$0.3&47.1$\pm$0.7&24.5$\pm$1.2&56.2$\pm$0.8&33.7$\pm$0.7&60.5&38.1     \\
          3DIoU~\cite{cvpr20213dioumatch}        &39.0$\pm$1.9&21.1$\pm$1.7&45.5$\pm$1.5&28.8$\pm$0.7&49.7$\pm$0.4&30.9$\pm$0.2&58.3$\pm$0.9&35.6$\pm$0.4&61.5&41.3    \\
          \textbf{Ours}&\textbf{41.1$\pm$1.2}&\textbf{21.8$\pm$1.8}&\textbf{47.4$\pm$0.8}&\textbf{29.2$\pm$1.2}&\textbf{53.4$\pm$0.9}&\textbf{31.2$\pm$1.3}&\textbf{60.1$\pm$0.4}&\textbf{37.8$\pm$0.8}&\textbf{62.7}&\textbf{42.1}\\
          \bottomrule
      \end{tabular}
  \end{center}
\end{table*}

For a fair comparison with existing methods~\cite{qi2019deep,cvpr2020sess,cvpr20213dioumatch}, we follow 3DIoUMatch~\cite{cvpr20213dioumatch} and train the proposed model on ScanNet and SUNRGB-D under different ratios of labeled data. 
Due to the randomness of the data splits, we report the $\mbox{mAP}_{25}$ and $\mbox{mAP}_{50}$ as mean $\pm$ standard deviation across 3 runs.
For 100\% labeled data, we make a copy of the full dataset as unlabeled data.
As shown in Table~\ref{table:scannet} and Table~\ref{table:sunrgbd}, 
our approach shows significant performance improvements on both ScanNet and SUNRGB-D benchmark and achieves new SOTA performance under different ratios of labeled data.
For ScanNet dataset, our method outperforms 3DIOUMatch by 1.8\% and 2.1\% for $\mbox{mAP}_{50}$ on 10\% and 20\% labeled datasets, respectively.
For SUNRGB-D dataset, our method outperforms 3DIOUMatch by 1.9\% and 3.7\% for $\mbox{mAP}_{50}$ on 10\% and 20\% labeled datasets, respectively.
In addition, with only 50\% labeled data, our method achieves better performance than the fully supervised baseline model on both ScanNet and SUNRGB-D.
We notice that our method achieves performance gain when using 100\% labeled data. 
The performance improvement may come from the suppression of dataset noise by the proposed semi-supervised framework. 
\begin{figure}[!t]
  \begin{center}
      \includegraphics[width=0.45\textwidth]{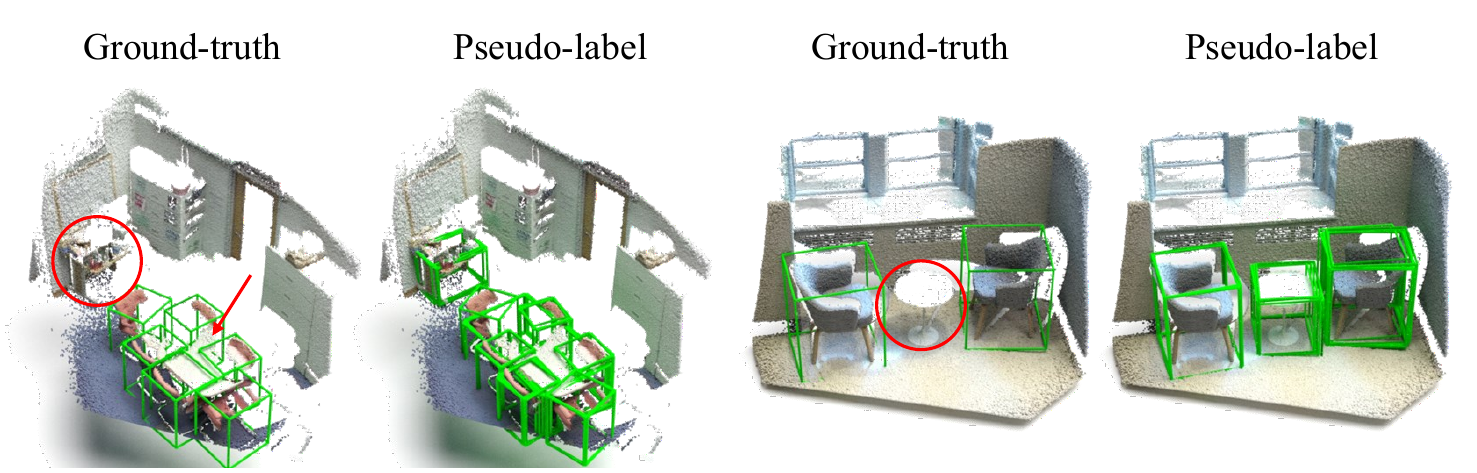}
      \caption{\textbf{Annotations with low-quality on SUNRGB-D dataset.} We train the model on 100\% labeled data and then generate pseudo-labels by the proposed soft-PLS.}
      \label{noisy_gt}
      \vspace{-2em}
  \end{center}
\end{figure}
As shown in Figure~\ref{noisy_gt}, the pseudo-label generated by the teacher model is a correction for missing or incorrect annotations, providing more valuable information for student model training.
\begin{table*}[!t]
    \begin{center}
        \setlength\tabcolsep{3.0pt}
        \caption{\textbf{Results on KITTI Val set under different ratios of labeled data.}
        The results are reported as mean $\pm$ standard deviation across 3 runs with random data splits.}
        \small
        \label{table:KITTI}
        \begin{tabular}{c|c|c|c|c|c|c|c|c|c|c|c|c}
            \toprule
            \multirow{2}*{Model} & \multicolumn{3}{c|}{mAP(1\%)} & \multicolumn{3}{c|}{mAP(10\%)} & \multicolumn{3}{c|}{mAP(20\%)} & \multicolumn{3}{c}{mAP(100\%)} \\
            \cline{2-13}
            ~& Car & Ped. & Cyc. & Car & Ped. & Cyc. & Car & Ped. & Cyc. & Car & Ped. & Cyc. \\
            \midrule
            PV-RCNN~\cite{shi2020pv}      &73.1$\pm$0.2&21.4$\pm$11.1&28.0$\pm$6.0&80.7$\pm$1.0&50.0$\pm$3.2&60.5$\pm$4.7&82.4$\pm$0.2&52.4$\pm$1.5&65.8$\pm$1.3&82.5&58.1&73.5\\
            3DIoU~\cite{cvpr20213dioumatch}   &75.2$\pm$1.8&32.9$\pm$16.1&31.4$\pm$7.8&81.3$\pm$0.8&52.6$\pm$1.9&62.0$\pm$5.8&82.9$\pm$0.1&54.5$\pm$1.4&67.4$\pm$1.7&84.2&60.5&75.2\\
            \textbf{Ours}&\textbf{76.3$\pm$1.0}&\textbf{33.1$\pm$13.6}&\textbf{33.6$\pm$5.2}&\textbf{83.1$\pm$0.5}&\textbf{54.2$\pm$1.9}&\textbf{65.3$\pm$3.6}&\textbf{84.1$\pm$0.2}&\textbf{57.8$\pm$1.5}&\textbf{70.8$\pm$0.5}&\textbf{85.3}&\textbf{60.9}&\textbf{76.3}\\
            \bottomrule
        \end{tabular}
    \end{center}
\end{table*}
\subsection{Comparison on Outdoor Dataset}
To verify the adaptability of our method to object detection of different scenes, we conduct experiments on KITTI benchmark. 
Similar to the indoor datasets, we train the proposed model under different ratios of labeled data and report the results as mean $\pm$ standard. 
Table~\ref{table:KITTI} shows the competitive results on KITTI benchmark. 
Our method achieves significant improvements in all three categories, especially pedestrian and bicycle, where our method achieves 3.2\% and 3.4\% performance gains on 20\% labeled data, respectively. 
We note that the standard deviations of pedestrians and bicycles are larger than cars and this is caused by the number of labeled samples for these categories being rather small.
In order to validate the performance of the proposed semi-supervised approach in the case of extreme lack of data, we validate our method on KIITI 1\% labeled data and the model performance is significantly improved, especially for cars and bicycles. 

\subsection{Transductive Learning Comparison}
\begin{table}[!t]
  \begin{center}
    \setlength\tabcolsep{2.pt}
    \caption{\textbf{Transductive learning results on ScanNet dataset under different ratios of labeled data.}
      Results are reported as mean across 3 runs with random data splits.}
    \small
    \label{table:Transduction}
    \begin{tabular}{c|c|c|c|c|c|c}
      \toprule
      \multirow{2}*{Model} & \multicolumn{2}{c|}{10\%} & \multicolumn{2}{c|}{20\%} & \multicolumn{2}{c}{50\%}                                                                                                 \\
      \cline{2-7}
      ~                            & $\mbox{mAP}_{25}$         & $\mbox{mAP}_{50}$        & $\mbox{mAP}_{25}$     & $\mbox{mAP}_{50}$     & $\mbox{mAP}_{25}$     & $\mbox{mAP}_{50}$     \\
      \midrule
      VoteNet~\cite{qi2019deep}                          & 43.5              & 24.5             & 51.3          & 32.7          & 61.6          & 41.2          \\
      SESS~\cite{cvpr2020sess}                             & 48.8              & 28.7             & 56.6          & 36.9          & 67.6          & 44.2          \\
      3DIoU~\cite{cvpr20213dioumatch}                           & 51.2              & 32.7             & 57.8          & 38.1          & 66.8          & 45.9          \\
      \textbf{Ours}            & \textbf{53.8}     & \textbf{34.6}    & \textbf{60.2} & \textbf{40.4} & \textbf{68.5} & \textbf{48.5} \\
      \bottomrule
    \end{tabular}
  \end{center}
\end{table}
Semi-supervised learning uses both labeled data and unlabeled data for model training.
There are two categories of semi-supervised learning, based on the type of testing data: inductive learning and transductive learning.
Inductive learning use the new unseen data as testing data, while transductive learning use the unlabeled data that are used  in the training stage as testing data.
To evaluate the effectiveness of our method in transductive learning, we conduct experiments on the ScanNet dataset under different ratios of labeled data.
As shown in Table~\ref{table:Transduction}, our method outperforms all previous methods under different ratios of labeled data, which verifies the effectiveness of the proposed method for transductive semi-supervised learning.
%

%
\subsection{Ablation Studies}
\noindent\textbf{Evaluation of the model with different designs.}
\begin{table}[!t]
  \begin{center}
    \setlength\tabcolsep{2.0pt}
    \caption{\textbf{Evaluation of model with different designs.} SBBP is the Side-aware Bounding Box Parameterization. Soft-PLS is the Soft Pseudo-Label Selection. UEM is the Uncertainty Estimation Module. $F_{geo}$ and $F_{dist}$ are the geometric and distribution properties, respectively.}
    \small
    \label{table:side_aware}
  \begin{tabular}{c|c|cc|c|c|c|c}
    \toprule
    \multirow{2}*{SBBP} & \multirow{2}*{Soft-PLS} & \multicolumn{2}{c|}{UEM} & \multicolumn{2}{c|}{ScanNet 20\%} & \multicolumn{2}{c}{ScanNet 50\%} \\
    \cline{5-8}
    ~&~& $F_{geo}$& $F_{dist}$& $\mbox{mAP}_{25}$ & $\mbox{mAP}_{50}$ & $\mbox{mAP}_{25}$ & $\mbox{mAP}_{50}$\\
    \midrule
    \ding{55}&\ding{55}&\ding{55}&\ding{55}&50.21&30.12&58.14&38.19\\
    \ding{51}&\ding{55}&\ding{55}&\ding{55}&50.38&30.42&58.17&38.43\\
    \ding{51}&\ding{51}&\ding{55}&\ding{55}&51.19&33.28&57.44&39.51\\
    \ding{51}&\ding{51}&\ding{51}&\ding{55}&52.56&35.15&60.26&41.82\\
    \ding{51}&\ding{51}&\ding{55}&\ding{51}&53.86&36.41&61.13&42.18\\
    \ding{51}&\ding{51}&\ding{51}&\ding{51}&\textbf{54.51}&\textbf{37.29}&\textbf{61.52}&\textbf{43.13}\\
    \bottomrule
\end{tabular}
\end{center}
\end{table}
In this section, we present our extensive ablation studies on the ScanNet dataset to evaluate the effectiveness of different designs.  
Table~\ref{table:side_aware} reports the performance of our model under various settings.
In the first line, we use VoteNet~\cite{qi2019deep} as the backbone and adopt a fixed threshold setting for pseudo-label selection. 
In the second line, we modify the baseline model with our proposed Side-aware Bounding Box Parameterization.
Although the parameterization approach does not yield a significant improvement in the model performance, it sets a foundation for subsequent designs.
In the third line, we utilize the distribution variance as the quality assessment of the sides and then filter the pseudo-labels by soft-PLS.
The performance improvement indicates the effectiveness of our proposed pseudo-label filtering strategy.
In the fourth line, we evaluate the uncertainty of the sides using geometric features extracted from the SoI pooling. 
In the fifth line, we use distribution properties for the same purpose. 
Both methods achieve better performance than the distribution variance. 
Finally, in the last line, we combine the geometric and distribution properties to evaluate the uncertainty of the sides and achieve the best results.
%

\noindent\textbf{Effect of the Parameterization Method.}
\begin{table}[!t]
  \begin{center}
    \setlength\tabcolsep{3.0pt}
    \caption{\textbf{Effect of the Parameterization method.} Naive parameterization is to predict the center and sizes of the 3D bounding boxes, while side-aware parameterization is to predict six sides of the 3D bounding boxes. Probabilistic method means predicting a probabilistic distribution for each regression value.}
    \small
    \label{table:Parameterization}
  \begin{tabular}{cc|c|c|c|c}
    \toprule
    \multirow{2}*{Method} & \multirow{2}*{Prob.} & \multicolumn{2}{c|}{ScanNet 20\%} & \multicolumn{2}{c}{ScanNet 50\%} \\
    \cline{3-6} 
    ~&~& $\mbox{mAP}_{25}$ & $\mbox{mAP}_{50}$ & $\mbox{mAP}_{25}$ & $\mbox{mAP}_{50}$\\
    \midrule
    w/ naive param. &\ding{55}&50.48&31.58&58.55&38.67\\
    w/ naive param. &\ding{51}&51.64&32.66&59.05&39.13\\
    w/ side. param.  &\ding{55}&52.21&35.02&60.07&41.29\\
    w/ side. param. &\ding{51}&\textbf{54.51}&\textbf{37.29}&\textbf{61.52}&\textbf{43.13}\\
    \bottomrule
\end{tabular}
\end{center}
\end{table}
To demonstrate the effectiveness of our proposed parameterization method for semi-supervised 3D object detection, we conduct detailed ablation experiments, as presented in Table~\ref{table:Parameterization}.
In the first line, we use a naive parameterization method to estimate the uncertainty of centers and sizes in the uncertainty estimation module. 
Additionally, we change the weight assignment of each side to match the weight assignment of the corresponding regression values.
In the second line, we make the parameterization approach probabilistic and use both distribution properties and geometric features to estimate the uncertainty of the regression values.
However, both the methods result in severe performance degradation due to the strong correlation between regression values in the naive parameterization method. 
This correlation make it difficult to accurately assess the localization quality of each regression values individually. 
The results from the third and fourth lines confirm the effectiveness of the distribution properties in the uncertainty estimation module.
%
%
%

\noindent\textbf{Effect of the Soft Pseudo-Label Selection.}
\begin{table}[!t]
  \begin{center}
    \setlength\tabcolsep{3.0pt}
    \caption{\textbf{Effect of the Soft Pseudo-Label Selection.} LHS is the IoU-guided NMS
    with Low-Half Strategy. CSF is the Category-Specific Filter with adaptive threshold. SWA is the Side-aware Weight Assignment. }
    \small
    \label{table:soft_PLS}
  \begin{tabular}{ccc|c|c|c|c}
    \toprule
    \multirow{2}*{LHS} & \multirow{2}*{CSF} & \multirow{2}*{SWA} & \multicolumn{2}{c|}{ScanNet 20\%} & \multicolumn{2}{c}{ScanNet 50\%} \\
    \cline{4-7}
    ~&~&~& $\mbox{mAP}_{25}$ & $\mbox{mAP}_{50}$ & $\mbox{mAP}_{25}$ & $\mbox{mAP}_{50}$\\
    \midrule
    \ding{55}&\ding{55}&\ding{55}&50.38&30.42&58.17&38.43\\
    \ding{51}&\ding{55}&\ding{55}&52.76&35.21&59.82&41.19\\
    \ding{51}&\ding{51}&\ding{55}&53.21&35.79&59.98&41.74\\
    \ding{51}&\ding{51}&\ding{51}&\textbf{54.51}&\textbf{37.29}&\textbf{61.52}&\textbf{43.13}\\
    \ding{55}&\ding{51}&\ding{51}&54.03&36.58&60.44&42.51\\
    \bottomrule
\end{tabular}
\end{center}
\end{table}
We present Table~\ref{table:soft_PLS} to demonstrate the impact of each designed component on the pseudo-label selection strategy.
In the first line, we utilize fixed thresholds to remove low-quality pseudo-labels.
As shown in the second line, we introduce the IoU-guided NMS with a low-half strategy to reduce the interference of repeated bounding boxes during model training, which  improves performance.
In the third line, we adopt category-specific adaptive thresholds to address the long-tail effects and retain more pseudo-labels for model training.
The model performance has achieved further improvement.
Finally, we ultize the side-aware weight assignment strategy based on the uncertainty of each side. 
This approach effectively suppresses the interference of false regression values on model training, resulting in a significant performance improvement.
To further demonstrate the effectiveness of the side-aware weight assignment strategy, we remove the IoU-guided NMS with a low-half strategy. 
The results indicate that the strategy can suppress the interference of noise in repeated bounding boxes, thereby maintaining good model performance.


\subsection{Sensitivity Analysis}
%
\begin{table}[!t]
  \begin{center}
      \setlength\tabcolsep{3.0pt}
      \caption{\textbf{Sensitivity analysis of the heuristics in the soft-PLS.}}
      \small
      \label{table:Sensitivity}
      \begin{tabular}{c|c|c|c|c|c|c}
          \toprule

          \multirow{2}*{$\tau_{obj}$} & \multirow{2}*{$\tau_{cls}^{min}$} &  \multirow{2}*{$\tau_{cls}^{max}$} &\multirow{2}*{$\tau_{IoU}^{min}$} &\multirow{2}*{$\tau_{IoU}^{max}$} & \multicolumn{2}{c}{ScanNet 20\%} \\
          \cline{6-7}
          ~&~& ~&~& ~& $\mbox{mAP}_{25}$ & $\mbox{mAP}_{50}$\\
          \midrule
          0.60 &0.50 & 0.80 &0.10 &0.20 &53.77 & 36.91\\
          0.70 &0.60 & 0.85 &0.15 &0.25 &\textbf{54.89} & 37.14\\
          0.80 &0.70 & 0.90 &0.15 &0.25 &54.51 & \textbf{37.29}\\
          0.90 &0.70 & 0.95 &0.20 &0.30 &53.15 & 36.79\\
          0.90 &0.80 & 0.95 &0.25 &0.35 &51.57 & 34.73\\
          \bottomrule
      \end{tabular}
  \end{center}
\end{table}
As shown in Table~\ref{table:Sensitivity}, we present the sensitivity analysis of the heuristics in the Soft Pseudo-Label Selection. 
By using different threshold settings to select pseudo-labels, we evaluate the robustness of our approach.
The performance of the model only slightly decreased when using lower thresholds, which further confirms the effectiveness of our proposed soft-PLS in suppressing the interference of low-quality pseudo-labels on model training.
However, setting higher thresholds results in a significant decrease in model performance. This is due to the fact that the high threshold filters out a large number of valuable pseudo-labels.

\section{Conclusion}
In this paper, we propose a side-aware framework with three specific designs: a probabilistic parameterization method, an uncertainty estimation module, and a soft pseudo-label selection.
To the best of our knowledge, our approach is the first to consider the quality of local sides for pseudo-label filtering, enabling full exploitation and utilization of valid information in the model prediction results for supervising student models.
Experiment results indicate that our method outperforms state-of-the-art methods on two indoor datasets and one outdoor dataset.

{\small
\bibliographystyle{ieee_fullname}
\bibliography{egbib}
}

\end{document}